\documentclass[12pt]{article}
\usepackage{amsmath}
\usepackage{graphicx}
\usepackage{hyperref}
\usepackage{natbib}
\usepackage{geometry}
\usepackage{fancyhdr}
\usepackage{multirow}
\usepackage{multicol}
\usepackage{array}
\usepackage{blindtext}
\usepackage{booktabs}
\usepackage{bbm}%为了写粗体的1，在3.2节公式2
\begin{document}

\title{Hear the Scene: Audio-Enhanced Text Spotting}

\author{Jing Li, Bo Wang
\\
Zhejiang Gongshang University
\\
jingli@mail.zjgsu.edu.cn
}
\date{}
\maketitle

\begin{abstract}
Recent advancements in scene text spotting have focused on end-to-end methodologies that heavily rely on precise location annotations, which are often costly and labor-intensive to procure. In this study, we introduce an innovative approach that leverages only transcription annotations for training text spotting models, substantially reducing the dependency on elaborate annotation processes. Our methodology employs a query-based paradigm that facilitates the learning of implicit location features through the interaction between text queries and image embeddings. These features are later refined during the text recognition phase using an attention activation map. Addressing the challenges associated with training a weakly-supervised model from scratch, we implement a circular curriculum learning strategy to enhance model convergence. Additionally, we introduce a coarse-to-fine cross-attention localization mechanism for more accurate text instance localization. Notably, our framework supports audio-based annotation, which significantly diminishes annotation time and provides an inclusive alternative for individuals with disabilities. Our approach achieves competitive performance against existing benchmarks, demonstrating that high accuracy in text spotting can be attained without extensive location annotations.

\end{abstract}

\begin{figure}[t]
\includegraphics[width=0.475\textwidth] {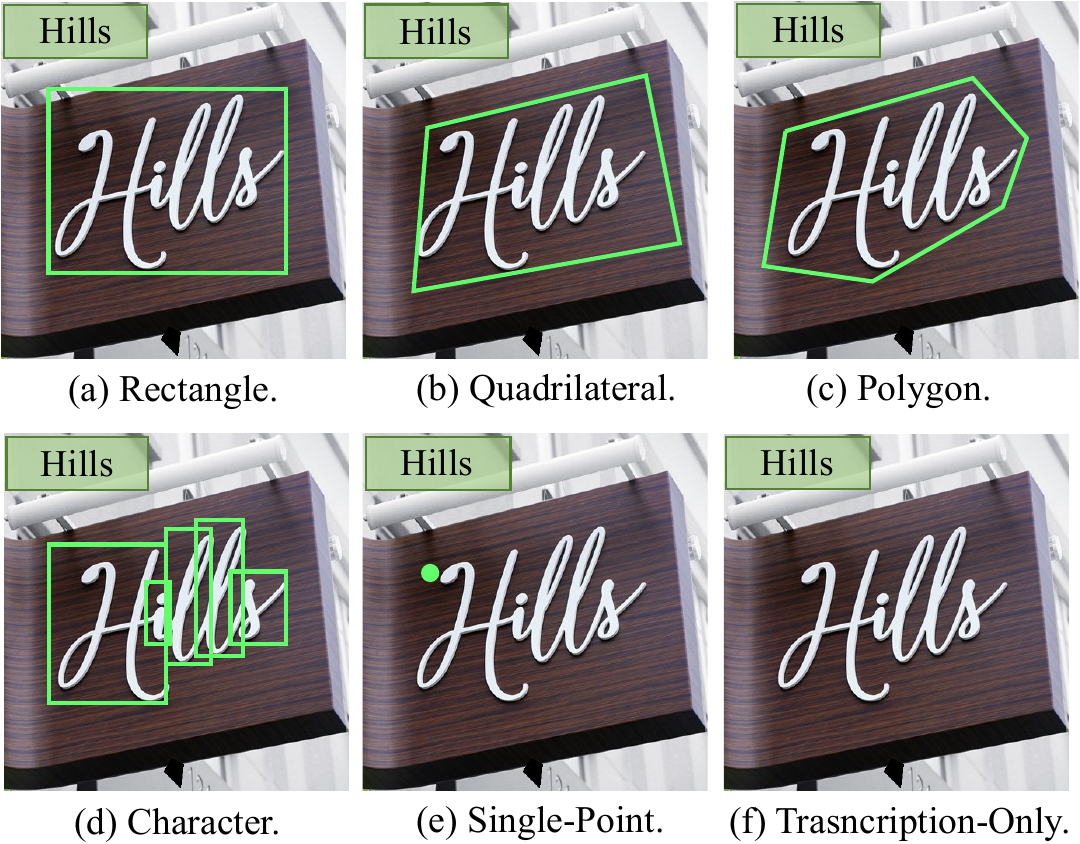}
  \centering
   \caption{Different annotation styles for text spotting. The blue point and lines are the position annotation for the text. The transcription annotation ``Hills'' is displayed in the top left corner of each image. 
   %Transcription-only annotation style does not have any location information of text.
   }
   \label{fig-annotation-style}
\end{figure}

\section{Introduction}\label{Sec:Intro}

Scene text spotting \cite{tang2022optimal, 2020ABCNet, Char-Net, liu2023spts, liu2020abcnet, tang2022few, tang2022youcan} remains a vibrant area of research due to its extensive applications across various domains such as image-based translation, multimedia content analysis, and accessibility features in visual media. The field \cite{MTS_V3, zhao2024multi, feng2023docpedia, feng2023unidocuniversallargemultimodal} has evolved from focusing primarily on horizontal or multi-oriented text to addressing more complex, arbitrarily-shaped text. This evolution has prompted shifts from traditional annotation formats like horizontal rectangles and quadrilaterals to more sophisticated polygonal annotations, as depicted in Figure 1.
Traditionally, scene text spotting involved separate processes for text detection and recognition, where both tasks required annotations for training. Text recognition used transcription annotations, and text detection depended on geometric annotations like polygons or bounding boxes. The reliance on detailed geometric annotations for training not only escalates the cost but also increases the manpower required, making the process less efficient and scalable.
Recent advancements have introduced methods \cite{li2024seed, li2024llava, zhao2024harmonizing, zhao2024tabpedia, tang2024mtvqabenchmarkingmultilingualtextcentric, tang2023character, tang2024textsquare} that directly identify and classify characters or words within texts, which, while reducing some dependencies, still rely heavily on character-level annotations and thus remain costly. Furthermore, the granularity of these annotations makes the task time-consuming and limits the feasibility of large-scale applications on real-world data. For instance, the annotation of text in videos or cluttered urban scenes can be prohibitively expensive and labor-intensive due to the complexity and volume of data.
Given these challenges, there is a growing interest in exploring methods that reduce reliance on expensive annotations. Preliminary efforts \cite{wang2024enhancing, wang2025mari, sun2025attentive}, such as the single-point annotation strategy proposed by SPTS \cite{liu2023spts}, have marked a significant step forward but still leave room for improvement. This paper introduces a novel approach that rethinks scene text spotting from an image captioning perspective, where the text's location is inferred implicitly through visual attention mechanisms, without direct location supervision. We propose a fully end-to-end architecture, termed  EchoSpot, which leverages a query-based interaction between text queries and image embeddings to approximate text locations while ensuring robust text recognition.
Our contributions are geared towards simplifying the annotation process, reducing costs, and making text spotting more accessible and scalable. By demonstrating that accurate text spotting can be achieved without explicit location annotations and by integrating novel strategies like circular curriculum learning and audio annotations, this work paves the way for future research and practical applications in the field of optical character recognition and beyond.

\begin{figure*}[t]
\includegraphics[width=0.9\linewidth] {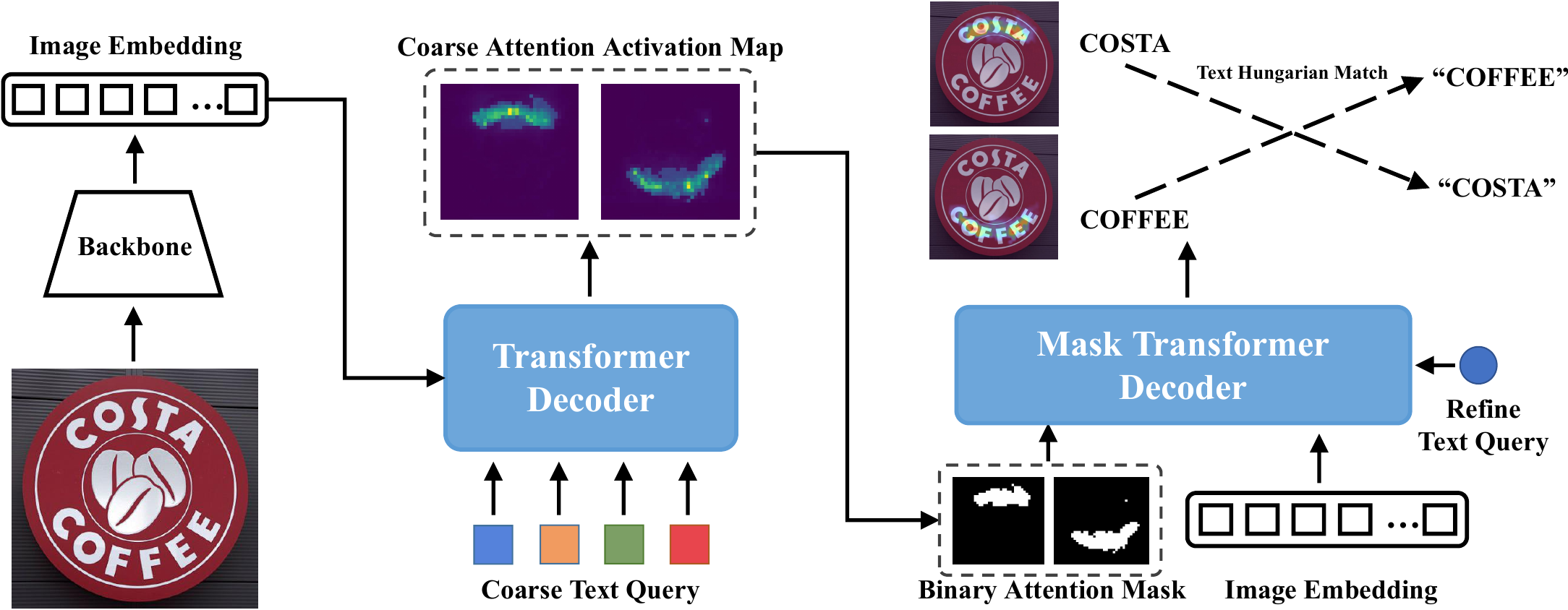}
  \centering
%   \vspace{-0.1cm}
   \caption{Overview of the proposed EchoSpot model architecture. 
   The visual and contextual embeddings are first extracted by a backbone network. The embeddings are then decoded to focus on the relevant text regions using a query-based cross-attention module. Next, the refine-stage text query, the mask generated by the attention activation map, and image embeddings are decoded together to obtain a refined text position.}
   \label{fig-network}
\end{figure*}

\section{Related Work}\label{Sec:RW}

The field of scene text spotting has witnessed a gradual shift from reliance on heavily annotated data towards methods that require fewer annotations. This evolution reflects the increasing need for cost-effective and scalable solutions. This section reviews existing literature, categorizing the studies into fully-supervised and weakly-supervised methods.
\subsection{Fully-Supervised Text Spotting}
Traditionally, text spotting techniques have drawn heavily from generic object detection frameworks, employing two-stage models that first detect text and then recognize it. Early approaches, like the one proposed by Li et al. \cite{crnn}, utilized CNN-based detectors coupled with RNN-based recognition branches, applying RoI pooling to crop text regions for subsequent recognition. These models, however, were limited to handling horizontally oriented text. Extensions of these methods introduced adaptations to manage text with arbitrary orientations and shapes. For example, He et al. \cite{Char-Net} expanded the RoIAlign method to extract features of text in arbitrary orientations, and Sun et al. \cite{2018FOTS} introduced a perspective RoI transform layer to align quadrangle proposals into feature maps. More recent innovations, such as the Mask TextSpotter series \cite{MTS_V3}, have advanced capabilities to handle arbitrarily-shaped text through character-level semantic segmentation, which, while effective, significantly increases the annotation burden due to the necessity for precise character localization.
\subsection{Weakly-Supervised Text Spotting}
The high cost and labor intensity of annotations required by fully-supervised methods 
\cite{wang2025pargo, feng2025docpedia, shan2024mctbenchmultimodalcognitiontextrich, sun2024attentiveeraserunleashingdiffusion} have spurred interest in weakly-supervised approaches. These methods aim to reduce or entirely eliminate the need for detailed geometric annotations. Notable among the emerging solutions are systems that use minimalistic annotations, such as single-point annotations proposed by SPTS \cite{liu2023spts}, which represent text positions with single points, reducing annotation complexity. Other exploratory works found in preprints like arXiv suggest innovative directions such as using image captions to train models that can identify text regions without explicit location data. These methods leverage advancements in deep learning, such as contrastive learning and transformer architectures, to intuit text locations from high-level descriptions or minimal guidance.
Despite these advancements, there remains a discernible gap between the fully-supervised and weakly-supervised approaches in terms of accuracy and reliability, particularly in complex real-world scenarios. Our work builds on these foundations, aiming to bridge this gap by proposing a transcription-only supervised method that simplifies the annotation process to just text transcriptions, eliminating the need for any geometric data. This approach not only aligns with the trend toward reducing annotation dependencies but also enhances the scalability and accessibility of text spotting technologies.

\section{Methodology}\label{Sec:method}

The methodology of our  EchoSpot introduces several innovative components designed to effectively spot text without reliance on location annotations. The approach leverages the interaction between text queries and image embeddings, refined through sophisticated neural network mechanisms. This section outlines the core components of our methodology.
\subsection{Query-based Cross Attention}
At the heart of EchoSpot is the Query-based Cross Attention module, which utilizes a transformer architecture similar to those found in natural language processing. The module begins with randomly initialized text queries which are processed through stacked layers of self-attention and cross-attention mechanisms. These mechanisms enable the model to generate embeddings that represent potential text locations based on contextual clues from the image, effectively allowing the model to "learn" where texts are likely to be located without explicit geometrical annotations.
\subsection{Coarse-to-Fine Cross Attention Localization Mechanism}
Building on the foundations laid by the initial query-based attention, the Coarse-to-Fine Cross Attention Localization Mechanism refines the text spotting process further. This two-pronged approach first filters through the rough text regions identified by the initial module to discard irrelevant features. It then employs a more refined cross-attention process that sharpens the focus on text regions, enhancing the model's accuracy in spotting text locations.
Text-based Hungarian Matching Loss
This component introduces a novel loss function tailored for our transcription-only supervision approach. It uses the Hungarian matching algorithm to align predicted text sequences with their corresponding ground truths during training. This alignment helps in backpropagating accurate error gradients that improve both the recognition and localization aspects of the model, thereby enhancing overall learning efficiency.
\subsection{Circular Curriculum Learning Strategy}
To address the challenges associated with training from weak annotations, we devise a Circular Curriculum Learning Strategy. This strategy schedules the training process to initially focus on simpler, clearer text instances before gradually introducing more complex images. This pedagogical approach helps the model develop a robust understanding of various text forms and contexts, thereby improving its generalizability and effectiveness on real-world data.
\subsection{Audio Annotation}
In an effort to simplify the annotation process further and make it accessible, we introduce an audio annotation system. This system allows annotators to speak the text present in images, which is then converted into textual annotations through speech recognition technologies. This method not only speeds up the annotation process but also opens up possibilities for persons with visual impairments to contribute to dataset creation.
\subsection{Inference Process}
During inference, the model employs the learned embeddings to predict text locations and content. The process begins with extracting features from the input image, followed by the application of the coarse-to-fine mechanism to pinpoint text regions. The predicted regions are then processed for text recognition and classification, providing the final spotting results.
Through these methodological innovations, EchoSpot advances the state of text spotting by reducing dependency on costly annotations while maintaining competitive accuracy and efficiency.

\section{Experiments}

To evaluate the effectiveness of the  EchoSpot, we conducted comprehensive experiments across various benchmarks. This section describes the datasets used, the evaluation protocols, implementation details, and a detailed discussion of the results, including an ablation study to validate specific components of our methodology.
\subsection{Datasets}

Our experiments were performed using several well-known scene text benchmarks:

ICDAR 2013 (IC13) \cite{ic13}: Primarily contains horizontally aligned text.

ICDAR 2015 (IC15)\cite{ic15}: Includes more challenging, multi-oriented text.

Total-Text: Features curved and highly irregular text shapes.

SCUT-CTW1500 \cite{ctw1500}: Contains curved and annotated text lines.

For training, apart from specific benchmark datasets, we utilized SynthText and COCO-Text to provide a diverse range of text instances, enhancing the model's ability to generalize across different text types and backgrounds.
\subsection{Evaluation Protocol}
We adopted two main metrics for evaluation:
Polygon Metric: Assesses the precision of the detected text regions against ground truth polygons, using Intersection over Union (IoU) as a measure.
Single-point Metric: Evaluates the accuracy of the model in predicting the central point of text instances, beneficial for comparing with methods that do not provide exact bounding boxes.
\subsection{Implementation Details}
The EchoSpot model was implemented using a modular architecture comprising several key components detailed in the Methodology section. We employed a ResNet-50 backbone augmented with additional convolutional blocks to extract robust features from input images. The model was trained using Adam optimizer with a learning rate strategy adapted to the circular curriculum learning approach, gradually decreasing as the training progressed through cycles of increasing complexity.
\subsection{Experimental Results and Ablation Study}
Our results demonstrate competitive performance across all benchmarks, with particularly strong results in datasets involving curved and irregular text, underscoring the model’s ability to handle complex text spotting scenarios.
The ablation study specifically highlighted the impact of the coarse-to-fine mechanism and the circular curriculum learning strategy. Removing the coarse-to-fine mechanism led to a noticeable drop in performance, validating its role in refining text localization. Similarly, replacing the circular curriculum learning with traditional training approaches resulted in slower convergence and reduced final accuracy, emphasizing its effectiveness in managing the learning process from simple to complex scenes.
\subsection{Discussion}
The experiments confirm that EchoSpot can achieve high levels of accuracy without relying on detailed geometric annotations, significantly reducing the annotation burden and costs associated with traditional text spotting methods. The model's robustness across diverse types of text and its ability to learn from minimal supervision suggest promising directions for further research and practical applications in optical character recognition and related fields.

\begin{figure*}[t]
\includegraphics[width=1.\textwidth] {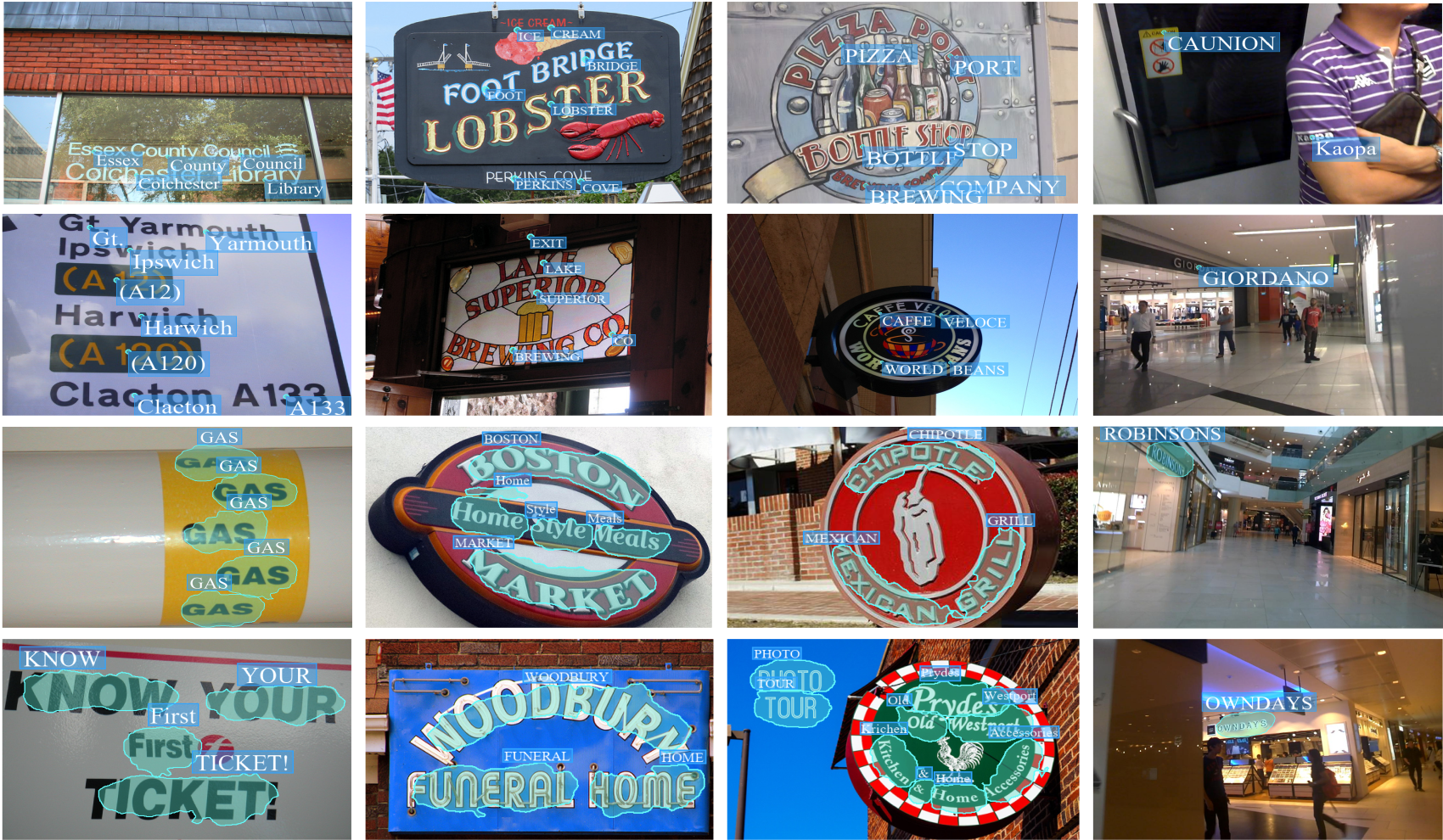}
  \centering
   \caption{Qualitative results. Images are selected from ICDAR 2013 (first col.), SCUT-CTW1500 (second col.), Total-Text (third col.), and ICDAR 2015 (fourth col.). 
   The first row contains visualizations of single-point, while the second row contains visualizations of masks. 
   %The first and second rows contain visualizations of single points, while the third and fourth rows contain visualizations of masks. 
   As shown in the figure, our method is robust against various  text types, including long text, large text, small text, curved text, perspective text, and fuzzy text. 
   }
   \label{fig-show-results}
\end{figure*}

\begin{table}[ht]
	\centering
	\renewcommand\arraystretch{1}
	\scalebox{1}{
		\begin{tabular}{c|p{1.3cm}<{\centering}|c}
			\hline
            Number of & \multicolumn{2}{c}{Total-Text End-to-End}\\
			\cline{2-3}
			 Coarse-to-Fine Attention & None & Full \\
			\hline
			0 & 58.5 & 67.4 \\
			1 & 65.1 & 74.8 \\
			2 & 66.3 & 75.2 \\
			\hline
	\end{tabular}}
	\caption{End-to-end recognition results on Total-Text w.r.t number of times of coarse-to-fine cross attention localization mechanism. ``None'' denotes lexicon-free. ``Full'' denotes that we use all the words appearing in the test set. All the results are obtained under the single-point metric.}
	\label{ab_refine_times}
\end{table}

\begin{table}[ht]
	\centering
	\renewcommand\arraystretch{1}
	\scalebox{1.0}{
		\begin{tabular}{r|c|c|c}
% 			\toprule
            \hline
			\multirow{2}{*}{Method} & \multicolumn{3}{c}{IC13 End-to-End}\\
			\cline{2-4}
			& S & W & G\\
			\hline
        	\multicolumn{4}{c}{Fully-supervised methods} \\
			\hline
			FOTS~\cite{2018FOTS} & 88.8 & 87.1 & 80.8 \\
			TextNet~\cite{sun2018textnet} & 89.8 & 88.9 & 83.0 \\
			Mask TextSpotter~\cite{Lyu_2018_ECCV} & \textbf{93.3} & \textbf{91.3} & \textbf{88.2} \\
			Boundary~\cite{Wang_Lu_Zhang_Yang_Bai_Xu_He_Wang_Liu_2020} & 88.2 & 87.7 & 84.1 \\
			Text Perceptron~\cite{TextPerceptron} & 91.4 & 90.7 & 85.8 \\
			\hline
			\multicolumn{4}{c}{Point-based methods} \\
			\hline
			SPTS (Single-Point)~\cite{peng2021spts} & 87.6 & 85.6 & 82.9 \\
			\hline
			\multicolumn{4}{c}{Transcription-only methods} \\
			\hline
			EchoSpot (Single-Point) (Ours) & 86.4 & 85.1 & 82.2\\
			EchoSpot (Polygon) (Ours) & 77.7 & 76.8 & 73.3 \\
			\hline
% 			\bottomrule
	\end{tabular}}
	\caption{End-to-end recognition results on ICDAR 2013. ``S'', ``W'', and ``G'' denote recognition with ``Strong'', ``Weak'', and ``Generic'' lexicon, respectively. ``Single Point'' and ``Polygon'' denote the two metrics mentioned in Sec.~\ref{sec-metric}. }
	\label{tab-ic13}
\end{table}

\begin{table}[ht]
	\centering
	\renewcommand\arraystretch{1}
	\scalebox{1}{
		\begin{tabular}{r|c|c|c}
			\hline
			\multirow{2}{*}{Method} & \multicolumn{3}{c}{IC15 End-to-End}\\
			\cline{2-4}
			& S & W & G \\
			\hline
			\multicolumn{4}{c}{Fully-Supervised methods} \\
			\hline
			FOTS~\cite{2018FOTS} & 81.1 & 75.9 & 60.8 \\
			Mask TextSpotter~\cite{Lyu_2018_ECCV} & 83.0 & 77.7 & 73.5\\
			CharNet~\cite{Char-Net} & 83.1 & \textbf{79.2} & 69.1\\
			TextDragon~\cite{Feng_2019_ICCV} & 82.5 & 78.3 & 65.2\\
			Mask TextSpotter v3~\cite{MTS_V3} & \textbf{83.3} & 78.1 & \textbf{74.2}\\
			MANGO~\cite{mango} & 81.8 & 78.9 & 67.3\\
			ABCNetV2~\cite{liu2020abcnet} & 82.7 & 78.5 & 73.0\\
			PAN++~\cite{9423611} & 82.7 & 78.2 & 69.2\\
			\hline
			\multicolumn{4}{c}{Point-based methods} \\
			\hline
			SPTS (Single-Point)~\cite{peng2021spts} & 64.6 & 58.8 & 54.9\\
			\hline
			\multicolumn{4}{c}{Transcription-only methods} \\
			\hline
			EchoSpot (Single-Point) (Ours) & 65.9 & 59.6 & 52.4\\
			EchoSpot (Polygon) (Ours)& 60.2 & 54.5 & 47.1 \\
			\hline
	\end{tabular}}
	\caption{End-to-end recognition results on ICDAR 2015. }
	\label{tab-ic15}
% 	\vspace{0.3cm}
\end{table}

\begin{table}[ht]
	\centering
	\renewcommand\arraystretch{1}
	\scalebox{1}{
		\begin{tabular}{r|p{1.3cm}<{\centering}|c}
			\hline
			\multirow{2}{*}{Method} & \multicolumn{2}{c}{Total-Text End-to-End}\\
			\cline{2-3}
			& None & Full \\
			\hline
			\multicolumn{3}{c}{Fully-Supervised methods} \\
			\hline
			CharNet~\cite{Char-Net} & 66.2 & -\\
			ABCNet~\cite{2020ABCNet} & 64.2 & 75.7\\
			PGNet~\cite{pgnet} & 63.1 & -\\
			Mask TextSpotter~\cite{Lyu_2018_ECCV} & 65.3 & 77.4 \\
			Qin et al~\cite{2019Towards} & 67.8 & - \\
			Mask TextSpotter v3~\cite{MTS_V3} & 71.2 & 78.4\\
			MANGO~\cite{mango} & \textbf{72.9} & \textbf{83.6} \\
			PAN++~\cite{9423611} & 68.6 & 78.6 \\
			ABCNet v2~\cite{2021ABCNet} & 70.4 & 78.1 \\
			\hline
			\multicolumn{3}{c}{Point-based methods} \\
			\hline
			SPTS (Single-Point)~\cite{peng2021spts} & 67.9 & 74.1\\
			\hline
			\multicolumn{3}{c}{Transcription-only methods} \\
			\hline
			EchoSpot (Single-Point) (Ours)& 65.1 &  74.8\\
			EchoSpot (Polygon) (Ours)& 61.5 & 73.0\\
			\bottomrule
	\end{tabular}}
	\caption{End-to-end recognition results on Total-Text. ``None'' denotes lexicon-free. ``Full'' denotes that we use all the words appearing in the test set.}
	\label{tab-totaltext}
% 	\vspace{0.3cm}
\end{table}

\begin{table}[ht]
	\centering
	\renewcommand\arraystretch{1} 
	\scalebox{1}{
		\begin{tabular}{r|p{1.7cm}<{\centering}|c}
			\hline
			\multirow{2}{*}{Method} & \multicolumn{2}{c}{SCUT-CTW1500 End-to-End}\\
			\cline{2-3}
			& None & Full \\
			\hline
			\multicolumn{3}{c}{Fully-Supervised methods} \\
			\hline
			TextDragon~\cite{Feng_2019_ICCV} & 39.7 & 72.4 \\
			ABCNet~\cite{2020ABCNet} & 45.2 & 74.1 \\
			MANGO~\cite{mango} & \textbf{58.9} & \textbf{78.7} \\
			ABCNet v2~\cite{2021ABCNet} & 57.5 & 77.2 \\
			\hline
			\multicolumn{3}{c}{Point-based methods} \\
			\hline
			SPTS (Single-Point)~\cite{peng2021spts} & 56.3 & 67.2 \\
			\hline
			\multicolumn{3}{c}{Transcription-only methods} \\
			\hline
			EchoSpot (Single-Point) (Ours) & 54.2 & 65.3 \\
			EchoSpot (Polygon) (Ours)& 51.4 & 61.7 \\ 
			\hline
	\end{tabular}}
	\caption{End-to-end recognition results on SCUT-CTW1500. }
	\label{tab-ctw1500}
% 	\vspace{0.3cm}
\end{table}

\begin{table}[h]
	\centering
	\renewcommand\arraystretch{1}
	\scalebox{1}{
		\begin{tabular}{r|p{1.3cm}<{\centering}|c}
			\hline
			\multirow{2}{*}{Annotation Style} & \multicolumn{2}{c}{Total-Text End-to-End}\\
			\cline{2-3}
			& None & Full \\
			\hline
			Text Transcription & 65.1 & 74.8 \\
			Audio (Word) & 64.3 & 73.9 \\
			Audio (Character) & 64.9 & 74.5 \\
			\hline
	\end{tabular}}
	\caption{End-to-end recognition results w.r.t annotation styles on Total-Text. ``Word'' and ``Character'' denote word-by-word and character-by-character pronunciation, respectively. 
	%``None'' denotes lexicon-free. ``Full'' denotes that we use all the words appeared in the test set.
	}
	\label{audio_ann_table}
\end{table}

\section{Conclusion}
In this paper, we introduced the  EchoSpot, a novel framework designed to address the challenges of scene text spotting with minimal reliance on costly and labor-intensive geometric annotations. By leveraging a query-based interaction between text queries and image embeddings, EchoSpot efficiently learns to spot text locations and recognize text content using only transcription annotations.
Our approach significantly simplifies the annotation process by allowing for text annotations through transcription and even audio inputs, making it not only cost-effective but also accessible to a broader range of contributors, including those with disabilities. The introduction of a circular curriculum learning strategy ensures that the model can be effectively trained from scratch on weakly annotated data, gradually learning to handle increasingly complex scenes through a structured learning approach.
The experimental results across several challenging datasets demonstrate that EchoSpot achieves competitive performance compared to state-of-the-art methods that rely on full supervision. Particularly, EchoSpot performs well on benchmarks involving curved and arbitrary-shaped text, underscoring its robustness and adaptability.
Future work will explore further enhancements to the coarse-to-fine localization mechanism to improve the precision of text localization. Additionally, extending the framework to support more languages and script types will increase its applicability on a global scale. We also see potential in exploring deeper integrations of audio-visual data to enrich the training process and improve the model’s performance in multimodal contexts.
Overall, the innovations presented in EchoSpot pave the way for more efficient and scalable text-spotting solutions, promising significant advancements in how optical character recognition technology is applied across various real-world applications.
\clearpage

\bibliographystyle{plainnat}
\bibliography{main}

\begin{thebibliography}{39}
\providecommand{\natexlab}[1]{#1}
\providecommand{\url}[1]{\texttt{#1}}
\expandafter\ifx\csname urlstyle\endcsname\relax
  \providecommand{\doi}[1]{doi: #1}\else
  \providecommand{\doi}{doi: \begingroup \urlstyle{rm}\Url}\fi

\bibitem[Feng et~al.(2023)Feng, Wang, Tang, Lu, Zhou, Li, and Huang]{feng2023unidocuniversallargemultimodal}
Hao Feng, Zijian Wang, Jingqun Tang, Jinghui Lu, Wengang Zhou, Houqiang Li, and Can Huang.
\newblock Unidoc: A universal large multimodal model for simultaneous text detection, recognition, spotting and understanding, 2023.
\newblock URL \url{https://arxiv.org/abs/2308.11592}.

\bibitem[Feng et~al.(2024)Feng, Liu, Liu, Jingqun, Zhou, Li, and Huang]{feng2025docpedia}
Hao Feng, Qi~Liu, Hao Liu, Tang Jingqun, Wengang Zhou, Houqiang Li, and Can Huang.
\newblock Docpedia: Unleashing the power of large multimodal model in the frequency domain for versatile document understanding.
\newblock \emph{SCIENCE CHINA Information Sciences}, 2024.

\bibitem[Feng et~al.(2019)Feng, He, Yin, Zhang, and Liu]{Feng_2019_ICCV}
Wei Feng, Wenhao He, Fei Yin, Xu-Yao Zhang, and Cheng-Lin Liu.
\newblock Textdragon: An end-to-end framework for arbitrary shaped text spotting.
\newblock In \emph{Proceedings of the IEEE/CVF International Conference on Computer Vision (ICCV)}, October 2019.

\bibitem[Karatzas et~al.(2013)Karatzas, Shafait, Uchida, Iwamura, i~Bigorda, Mestre, Mas, Mota, Almazan, and De~Las~Heras]{ic13}
Dimosthenis Karatzas, Faisal Shafait, Seiichi Uchida, Masakazu Iwamura, Lluis~Gomez i~Bigorda, Sergi~Robles Mestre, Joan Mas, David~Fernandez Mota, Jon~Almazan Almazan, and Lluis~Pere De~Las~Heras.
\newblock Icdar 2013 robust reading competition.
\newblock In \emph{Proc. ICDAR}, pages 1484--1493, 2013.

\bibitem[Karatzas et~al.(2015)Karatzas, Gomez-Bigorda, Nicolaou, Ghosh, Bagdanov, Iwamura, Matas, Neumann, Chandrasekhar, Lu, et~al.]{ic15}
Dimosthenis Karatzas, Lluis Gomez-Bigorda, Anguelos Nicolaou, Suman Ghosh, Andrew Bagdanov, Masakazu Iwamura, Jiri Matas, Lukas Neumann, Vijay~Ramaseshan Chandrasekhar, Shijian Lu, et~al.
\newblock Icdar 2015 competition on robust reading.
\newblock In \emph{ICDAR}, pages 1156--1160, 2015.

\bibitem[Li et~al.(2024{\natexlab{a}})Li, Zhang, Guo, Zhang, Li, Zhang, Zhang, Li, Liu, and Li]{li2024llava}
Bo~Li, Yuanhan Zhang, Dong Guo, Renrui Zhang, Feng Li, Hao Zhang, Kaichen Zhang, Yanwei Li, Ziwei Liu, and Chunyuan Li.
\newblock {Llava-onevision: Easy visual task transfer}.
\newblock \emph{arXiv preprint arXiv:2408.03326}, 2024{\natexlab{a}}.

\bibitem[Li et~al.(2024{\natexlab{b}})Li, Ge, Chen, Ge, Zhang, and Shan]{li2024seed}
Bohao Li, Yuying Ge, Yi~Chen, Yixiao Ge, Ruimao Zhang, and Ying Shan.
\newblock {Seed-bench-2-plus: Benchmarking multimodal large language models with text-rich visual comprehension}.
\newblock \emph{arXiv preprint arXiv:2404.16790}, 2024{\natexlab{b}}.

\bibitem[Liao et~al.(2020)Liao, Pang, Huang, Hassner, and Bai]{MTS_V3}
Minghui Liao, Guan Pang, Jing Huang, Tal Hassner, and Xiang Bai.
\newblock Mask textspotter v3: Segmentation proposal network for robust scene text spotting.
\newblock In Andrea Vedaldi, Horst Bischof, Thomas Brox, and Jan-Michael Frahm, editors, \emph{Computer Vision -- ECCV 2020}, pages 706--722, Cham, 2020. Springer International Publishing.
\newblock ISBN 978-3-030-58621-8.

\bibitem[Liu et~al.(2018)Liu, Ding, Shi, Chen, and Yan]{2018FOTS}
X.~Liu, L.~Ding, Y.~Shi, D.~Chen, and J.~Yan.
\newblock Fots: Fast oriented text spotting with a unified network.
\newblock In \emph{2018 IEEE/CVF Conference on Computer Vision and Pattern Recognition (CVPR)}, 2018.

\bibitem[Liu et~al.(2020{\natexlab{a}})Liu, Chen, Shen, He, and Wang]{2020ABCNet}
Y.~Liu, H.~Chen, C.~Shen, T.~He, and L.~Wang.
\newblock Abcnet: Real-time scene text spotting with adaptive bezier-curve network.
\newblock In \emph{2020 IEEE/CVF Conference on Computer Vision and Pattern Recognition (CVPR)}, 2020{\natexlab{a}}.

\bibitem[Liu et~al.(2020{\natexlab{b}})Liu, Chen, Shen, He, Jin, and Wang]{liu2020abcnet}
Yuliang Liu, Hao Chen, Chunhua Shen, Tong He, Lianwen Jin, and Liangwei Wang.
\newblock Abcnet: Real-time scene text spotting with adaptive bezier-curve network.
\newblock In \emph{proceedings of the IEEE/CVF conference on computer vision and pattern recognition}, pages 9809--9818, 2020{\natexlab{b}}.

\bibitem[Liu et~al.(2023)Liu, Zhang, Peng, Huang, Wang, Tang, Huang, Lin, Shen, Bai, et~al.]{liu2023spts}
Yuliang Liu, Jiaxin Zhang, Dezhi Peng, Mingxin Huang, Xinyu Wang, Jingqun Tang, Can Huang, Dahua Lin, Chunhua Shen, Xiang Bai, et~al.
\newblock Spts v2: single-point scene text spotting.
\newblock \emph{IEEE Transactions on Pattern Analysis and Machine Intelligence}, 2023.

\bibitem[Lyu et~al.(2018)Lyu, Liao, Yao, Wu, and Bai]{Lyu_2018_ECCV}
Pengyuan Lyu, Minghui Liao, Cong Yao, Wenhao Wu, and Xiang Bai.
\newblock Mask textspotter: An end-to-end trainable neural network for spotting text with arbitrary shapes.
\newblock In \emph{Proceedings of the European Conference on Computer Vision (ECCV)}, September 2018.

\bibitem[Peng et~al.(2021)Peng, Wang, Liu, Zhang, Huang, Lai, Zhu, Li, Lin, Shen, et~al.]{peng2021spts}
Dezhi Peng, Xinyu Wang, Yuliang Liu, Jiaxin Zhang, Mingxin Huang, Songxuan Lai, Shenggao Zhu, Jing Li, Dahua Lin, Chunhua Shen, et~al.
\newblock Spts: Single-point text spotting.
\newblock \emph{arXiv preprint arXiv:2112.07917}, 2021.

\bibitem[Qiao et~al.(2020)Qiao, Tang, Cheng, Xu, Niu, Pu, and Wu]{TextPerceptron}
Liang Qiao, Sanli Tang, Zhanzhan Cheng, Yunlu Xu, Yi~Niu, Shiliang Pu, and Fei Wu.
\newblock Text perceptron: Towards end-to-end arbitrary-shaped text spotting.
\newblock In \emph{Proceedings of the AAAI Conference on Artificial Intelligence}, volume~34, pages 11899--11907, 2020.

\bibitem[Qiao et~al.(2021)Qiao, Chen, Cheng, Xu, Niu, Pu, and Wu]{mango}
Liang Qiao, Ying Chen, Zhanzhan Cheng, Yunlu Xu, Yi~Niu, Shiliang Pu, and Fei Wu.
\newblock Mango: A mask attention guided one-stage scene text spotter.
\newblock In \emph{National Conference on Artificial Intelligence}, 2021.

\bibitem[Qin et~al.(2019)Qin, BiSsAcco, Raptis, Fujii, and Xiao]{2019Towards}
S.~Qin, A.~BiSsAcco, M.~Raptis, Y.~Fujii, and Y.~Xiao.
\newblock Towards unconstrained end-to-end text spotting.
\newblock \emph{IEEE}, 2019.

\bibitem[Shan et~al.(2024)Shan, Fei, Shi, Wang, Tang, Liao, Tang, Bai, and Huang]{shan2024mctbenchmultimodalcognitiontextrich}
Bin Shan, Xiang Fei, Wei Shi, An-Lan Wang, Guozhi Tang, Lei Liao, Jingqun Tang, Xiang Bai, and Can Huang.
\newblock Mctbench: Multimodal cognition towards text-rich visual scenes benchmark, 2024.
\newblock URL \url{https://arxiv.org/abs/2410.11538}.

\bibitem[Shi et~al.(2016)Shi, Xiang, and Cong]{crnn}
B.~Shi, B.~Xiang, and Y.~Cong.
\newblock An end-to-end trainable neural network for image-based sequence recognition and its application to scene text recognition.
\newblock \emph{IEEE Transactions on Pattern Analysis \& Machine Intelligence}, 39\penalty0 (11):\penalty0 2298--2304, 2016.

\bibitem[Sun et~al.(2024)Sun, Cui, Dong, and Tang]{sun2024attentiveeraserunleashingdiffusion}
Wenhao Sun, Benlei Cui, Xue-Mei Dong, and Jingqun Tang.
\newblock Attentive eraser: Unleashing diffusion model's object removal potential via self-attention redirection guidance, 2024.
\newblock URL \url{https://arxiv.org/abs/2412.12974}.

\bibitem[Sun et~al.(2025)Sun, Dong, Cui, and Tang]{sun2025attentive}
Wenhao Sun, Xue-Mei Dong, Benlei Cui, and Jingqun Tang.
\newblock Attentive eraser: Unleashing diffusion model’s object removal potential via self-attention redirection guidance.
\newblock In \emph{Proceedings of the AAAI Conference on Artificial Intelligence}, volume~39, pages 20734--20742, 2025.

\bibitem[Sun et~al.(2018)Sun, Zhang, Huang, Liu, Han, and Ding]{sun2018textnet}
Yipeng Sun, Chengquan Zhang, Zuming Huang, Jiaming Liu, Junyu Han, and Errui Ding.
\newblock Textnet: Irregular text reading from images with an end-to-end trainable network.
\newblock In \emph{Asian Conference on Computer Vision}, pages 83--99. Springer, 2018.

\bibitem[Tang et~al.(2022{\natexlab{a}})Tang, Qian, Song, Dong, Li, and Bai]{tang2022optimal}
Jingqun Tang, Wenming Qian, Luchuan Song, Xiena Dong, Lan Li, and Xiang Bai.
\newblock Optimal boxes: boosting end-to-end scene text recognition by adjusting annotated bounding boxes via reinforcement learning.
\newblock In \emph{European Conference on Computer Vision}, pages 233--248. Springer, 2022{\natexlab{a}}.

\bibitem[Tang et~al.(2022{\natexlab{b}})Tang, Qiao, Cui, Ma, Zhang, and Kanoulas]{tang2022youcan}
Jingqun Tang, Su~Qiao, Benlei Cui, Yuhang Ma, Sheng Zhang, and Dimitrios Kanoulas.
\newblock You can even annotate text with voice: Transcription-only-supervised text spotting.
\newblock In \emph{Proceedings of the 30th ACM International Conference on Multimedia}, MM '22, page 4154–4163, New York, NY, USA, 2022{\natexlab{b}}. Association for Computing Machinery.
\newblock ISBN 9781450392037.
\newblock \doi{10.1145/3503161.3547787}.
\newblock URL \url{https://doi.org/10.1145/3503161.3547787}.

\bibitem[Tang et~al.(2022{\natexlab{c}})Tang, Zhang, Liu, Yang, Jiang, Hu, and Bai]{tang2022few}
Jingqun Tang, Wenqing Zhang, Hongye Liu, MingKun Yang, Bo~Jiang, Guanglong Hu, and Xiang Bai.
\newblock Few could be better than all: Feature sampling and grouping for scene text detection.
\newblock In \emph{Proceedings of the IEEE/CVF Conference on Computer Vision and Pattern Recognition}, pages 4563--4572, 2022{\natexlab{c}}.

\bibitem[Tang et~al.(2023)Tang, Du, Wang, Zhou, Mei, Xue, Xu, and Zhang]{tang2023character}
Jingqun Tang, Weidong Du, Bin Wang, Wenyang Zhou, Shuqi Mei, Tao Xue, Xing Xu, and Hai Zhang.
\newblock Character recognition competition for street view shop signs.
\newblock \emph{National Science Review}, 10\penalty0 (6):\penalty0 nwad141, 2023.

\bibitem[Tang et~al.(2024{\natexlab{a}})Tang, Lin, Zhao, Wei, Wu, Liu, Feng, Li, Wang, Liao, et~al.]{tang2024textsquare}
Jingqun Tang, Chunhui Lin, Zhen Zhao, Shu Wei, Binghong Wu, Qi~Liu, Hao Feng, Yang Li, Siqi Wang, Lei Liao, et~al.
\newblock Textsquare: Scaling up text-centric visual instruction tuning.
\newblock \emph{arXiv preprint arXiv:2404.12803}, 2024{\natexlab{a}}.

\bibitem[Tang et~al.(2024{\natexlab{b}})Tang, Liu, Ye, Lu, Wei, Lin, Li, Mahmood, Feng, Zhao, Wang, Liu, Liu, Bai, and Huang]{tang2024mtvqabenchmarkingmultilingualtextcentric}
Jingqun Tang, Qi~Liu, Yongjie Ye, Jinghui Lu, Shu Wei, Chunhui Lin, Wanqing Li, Mohamad Fitri Faiz~Bin Mahmood, Hao Feng, Zhen Zhao, Yanjie Wang, Yuliang Liu, Hao Liu, Xiang Bai, and Can Huang.
\newblock Mtvqa: Benchmarking multilingual text-centric visual question answering, 2024{\natexlab{b}}.
\newblock URL \url{https://arxiv.org/abs/2405.11985}.

\bibitem[Wang et~al.(2025{\natexlab{a}})Wang, Shan, Shi, Lin, Fei, Tang, Liao, Tang, Huang, and Zheng]{wang2025pargo}
An-Lan Wang, Bin Shan, Wei Shi, Kun-Yu Lin, Xiang Fei, Guozhi Tang, Lei Liao, Jingqun Tang, Can Huang, and Wei-Shi Zheng.
\newblock Pargo: Bridging vision-language with partial and global views.
\newblock In \emph{Proceedings of the AAAI Conference on Artificial Intelligence}, volume~39, pages 7491--7499, 2025{\natexlab{a}}.

\bibitem[Wang et~al.(2020)Wang, Lu, Zhang, Yang, Bai, Xu, He, Wang, and Liu]{Wang_Lu_Zhang_Yang_Bai_Xu_He_Wang_Liu_2020}
Hao Wang, Pu~Lu, Hui Zhang, Mingkun Yang, Xiang Bai, Yongchao Xu, Mengchao He, Yongpan Wang, and Wenyu Liu.
\newblock All you need is boundary: Toward arbitrary-shaped text spotting.
\newblock \emph{Proceedings of the AAAI Conference on Artificial Intelligence}, 34\penalty0 (07):\penalty0 12160--12167, Apr. 2020.
\newblock \doi{10.1609/aaai.v34i07.6896}.
\newblock URL \url{https://ojs.aaai.org/index.php/AAAI/article/view/6896}.

\bibitem[Wang et~al.(2025{\natexlab{b}})Wang, Yang, He, Zhang, Chen, and Huang]{wang2025mari}
Jianhui Wang, Zhifei Yang, Yangfan He, Huixiong Zhang, Yuxuan Chen, and Jingwei Huang.
\newblock Mari: Material retrieval integration across domains.
\newblock \emph{arXiv preprint arXiv:2503.08111}, 2025{\natexlab{b}}.

\bibitem[Wang et~al.(2024)Wang, Zhang, He, Song, Shi, Li, Xu, Wu, Qian, Chen, et~al.]{wang2024enhancing}
Junqiao Wang, Zeng Zhang, Yangfan He, Yuyang Song, Tianyu Shi, Yuchen Li, Hengyuan Xu, Kunyu Wu, Guangwu Qian, Qiuwu Chen, et~al.
\newblock Enhancing code llms with reinforcement learning in code generation.
\newblock \emph{arXiv preprint arXiv:2412.20367}, 2024.

\bibitem[Wang et~al.(2021{\natexlab{a}})Wang, Zhang, Qi, Liu, Zhang, Lyu, Han, Liu, Ding, and Shi]{pgnet}
Pengfei Wang, Chengquan Zhang, Fei Qi, Shanshan Liu, Xiaoqiang Zhang, Pengyuan Lyu, Junyu Han, Jingtuo Liu, Errui Ding, and Guangming Shi.
\newblock Pgnet: Real-time arbitrarily-shaped text spotting with point gathering network.
\newblock \emph{AAAI. AAAI}, pages 2782--2790, 2021{\natexlab{a}}.

\bibitem[Wang et~al.(2021{\natexlab{b}})Wang, Xie, Li, Liu, Liang, Zhibo, Lu, and Shen]{9423611}
Wenhai Wang, Enze Xie, Xiang Li, Xuebo Liu, Ding Liang, Yang Zhibo, Tong Lu, and Chunhua Shen.
\newblock Pan++: Towards efficient and accurate end-to-end spotting of arbitrarily-shaped text.
\newblock \emph{IEEE Transactions on Pattern Analysis and Machine Intelligence}, pages 1--1, 2021{\natexlab{b}}.
\newblock \doi{10.1109/TPAMI.2021.3077555}.

\bibitem[Xing et~al.(2019)Xing, Tian, Huang, and Scott]{Char-Net}
Linjie Xing, Zhi Tian, Weilin Huang, and Matthew~R Scott.
\newblock Convolutional character networks.
\newblock In \emph{Proc. ICCV}, pages 9126--9136, 2019.

\bibitem[Yuliang et~al.(2017)Yuliang, Lianwen, Shuaitao, and Sheng]{ctw1500}
Liu Yuliang, Jin Lianwen, Zhang Shuaitao, and Zhang Sheng.
\newblock Detecting curve text in the wild: New dataset and new solution.
\newblock \emph{arXiv preprint arXiv:1712.02170}, 2017.

\bibitem[Zhao et~al.(2024{\natexlab{a}})Zhao, Feng, Liu, Tang, Wu, Liao, Wei, Ye, Liu, Zhou, et~al.]{zhao2024tabpedia}
Weichao Zhao, Hao Feng, Qi~Liu, Jingqun Tang, Binghong Wu, Lei Liao, Shu Wei, Yongjie Ye, Hao Liu, Wengang Zhou, et~al.
\newblock Tabpedia: Towards comprehensive visual table understanding with concept synergy.
\newblock \emph{Advances in Neural Information Processing Systems}, 37:\penalty0 7185--7212, 2024{\natexlab{a}}.

\bibitem[Zhao et~al.(2024{\natexlab{b}})Zhao, Tang, Lin, Wu, Huang, Liu, Tan, Zhang, and Xie]{zhao2024multi}
Zhen Zhao, Jingqun Tang, Chunhui Lin, Binghong Wu, Can Huang, Hao Liu, Xin Tan, Zhizhong Zhang, and Yuan Xie.
\newblock Multi-modal in-context learning makes an ego-evolving scene text recognizer.
\newblock In \emph{Proceedings of the IEEE/CVF Conference on Computer Vision and Pattern Recognition}, pages 15567--15576, 2024{\natexlab{b}}.

\bibitem[Zhao et~al.(2024{\natexlab{c}})Zhao, Tang, Wu, Lin, Wei, Liu, Tan, Zhang, Huang, and Xie]{zhao2024harmonizing}
Zhen Zhao, Jingqun Tang, Binghong Wu, Chunhui Lin, Shu Wei, Hao Liu, Xin Tan, Zhizhong Zhang, Can Huang, and Yuan Xie.
\newblock Harmonizing visual text comprehension and generation.
\newblock \emph{arXiv preprint arXiv:2407.16364}, 2024{\natexlab{c}}.

\end{thebibliography}

\end{document}